\newif\ifcomments
\commentsfalse

\documentclass{article}
\PassOptionsToPackage{square, numbers, compress,sort}{natbib}
\usepackage[final]{nips_2017}

\usepackage[resetlabels]{multibib}
\usepackage[T1]{fontenc}    
\usepackage{booktabs}       
\usepackage{amsfonts}       
\usepackage{nicefrac}       
\usepackage{microtype}      
\usepackage[utf8]{inputenc}
\usepackage{color}
\usepackage{graphicx}
\usepackage{subcaption}
\usepackage{algorithm,algorithmic}
\usepackage{subcaption}
\usepackage{aliascnt}
\usepackage{hyperref}
\usepackage{url}            
\usepackage{amsfonts}
\usepackage{amsmath}
\usepackage{amsthm}
\usepackage{pgfplots}
\usepackage{booktabs}
\usepackage{wrapfig}
\usepackage[short]{optidef}
\usepackage[inline]{enumitem}
\usepgfplotslibrary{groupplots}

\newlist{inlinelist}{enumerate*}{1}
\setlist*[inlinelist,1]{%
  label=\arabic*),
}

\newcommand{\titl}{On Fairness and Calibration }
\newcommand{\authorinfo}{
  Geoff Pleiss\thanks{Equal contribution, alphebetical order.}, \:
  Manish Raghavan\footnotemark[1], \:
  Felix Wu, \:
  Jon Kleinberg, \:
  Kilian Q. Weinberger
  \\
  Cornell University, Department of Computer Science
  \\
  \texttt{\{geoff,manish,kleinber\}@cs.cornell.edu},
  \\
  \texttt{\{fw245,kwq4\}@cornell.edu}
}

\title{\titl}
\author{\authorinfo}

\ifcomments\newcommand{\comments}[1]{#1}\else\newcommand{\comments}[1]{}\fi
\definecolor{clrgp}{rgb}{.9,0,.9}

\newtheorem{definition}{Definition}
\newtheorem*{thm*}{Theorem}
\newtheorem{thm}{Theorem}
\newtheorem{corollary}{Corollary}[thm]
\newaliascnt{lemma}{thm}
\newtheorem{lemma}[lemma]{Lemma}
\aliascntresetthe{lemma}

\newcommand\numberthis{\addtocounter{equation}{1}\tag{\theequation}}

\renewcommand{\paragraph}[1]{

\textbf{#1}}

\newcommand{\R}{\mathbb{R}}

\newcommand{\x}{\mathbf{x}}

\newcommand{\ev}{\mathop{\mathbb{E}}}

\let\Pr\relax
\DeclareMathOperator*{\Pr}{P}

\DeclareMathOperator*{\argmin}{argmin}
\DeclareMathOperator*{\argmax}{argmax}

\newcommand{\hardtetal}{EO-Derived}
\newcommand{\zafaretal}{EO-Trained}

\newcommand{\fair}[1]{\tilde{#1}}
\newcommand{\eo}[1]{{#1}^{eo}}

\newcommand{\br}{\mu}

\newcommand{\fp}[1]{\ensuremath{c_{fp} ( #1 )}}
\newcommand{\fn}[1]{\ensuremath{c_{fn} ( #1 )}}

\newcommand{\calib}[1]{\ensuremath{\epsilon ( #1 )}}

\newcommand{\fairconst}[2]{\ensuremath{g_{#2}(#1)}}
\newcommand{\fairconstpr}[2]{\ensuremath{g'_{#2}(#1)}}
\newcommand{\fairconstname}{\ensuremath{g_{t}}}

\begin{document}

\maketitle
\setcounter{footnote}{0}

\begin{abstract}
The machine learning community has become increasingly concerned with the
potential for bias and discrimination in predictive models.
This has motivated a growing line of work on what it means for a classification procedure
to be ``fair.''
In this paper, we investigate the tension between minimizing error
disparity across different population groups while maintaining calibrated probability
estimates.
We show that calibration is compatible only with a single error
constraint (i.e. equal false-negatives rates across groups),
and show that any algorithm that satisfies this relaxation is no
better than randomizing a percentage of predictions for an existing classifier.
These unsettling findings, which extend and generalize existing results,
are empirically confirmed on several datasets.

\end{abstract}

\section{Introduction}

Recently, there has been growing concern about errors of machine learning algorithms in sensitive domains -- including criminal justice, online advertising, and
medical testing \cite{whitehouse-big-data} -- which may systematically discriminate against
particular groups of people \cite{barocas2016big,berk2017fairness,chouldechova2017fair}.
%
%
A recent high-profile example of these concerns was raised by the
news organization ProPublica, who studied a risk-assessment tool
that is widely used in the criminal justice system.
This tool assigns to each criminal defendant an estimated probability
that they will commit a future crime.
ProPublica found that the risk estimates assigned to
defendants who did not commit future crimes were on average
higher among African-American defendants than Caucasian defendants \cite{angwin-propublica-risk-scores}.
This is a form of false-positive error, and in this case it
disproportionately affected African-American defendants.
To mitigate issues such as these,
the machine learning community has proposed different
frameworks that attempt to quantify fairness in classification~\cite{barocas2016big,berk2017fairness,chouldechova2017fair,hardt2016equality,kleinberg2016inherent,woodworth2017learning,zafar2017fairness}.
A recent and particularly noteworthy framework is Equalized Odds \cite{hardt2016equality}
(also referred to as Disparate Mistreatment \cite{zafar2017fairness}),\footnote{
  For the remainder of the paper, we will use \emph{Equalized Odds}
  to refer to this notion of non-discrimination.
}
which constrains classification algorithms such that no error type (false-positive
or false-negative) disproportionately affects any population subgroup.
This notion of non-discrimination is feasible in many settings,
and researchers have developed tractable algorithms for achieving it \cite{hardt2016equality,goh2016satisfying,zafar2017fairness,woodworth2017learning}.
%

When risk tools are used in practice, a key goal is that they are
{\em calibrated}: if we look at the set of people who receive a predicted probability
of $p$, we would like a $p$ fraction of the members of this set to
be positive instances of the classification problem \cite{dawid1982well}.
Moreover, if we are concerned about fairness between two groups
$G_1$ and $G_2$ (e.g. African-American defendants and white defendants)
then we would like this calibration condition
to hold simultaneously for the set
of people within each of these groups as well
\cite{flores-re-propublica-fair}.
%
Calibration is a crucial condition for risk tools in many settings.
If a risk tool for
evaluating defendants were not calibrated with respect to groups
defined by race, for example, then a probability estimate of $p$
could carry different meaning for African-American and white defendants,
and hence the tool would have the unintended and highly undesirable
consequence of incentivizing judges to take race into account when
interpreting its predictions.
%
Despite the importance of calibration as a property,
our understanding of how it interacts
with other fairness properties is limited.
We know from recent work that, except in the most constrained
cases, it is impossible to achieve calibration while also
satisfying Equalized Odds
\citep{kleinberg2016inherent,chouldechova2017fair}.
However, we do not know how best to achieve relaxations of these guarantees
that are feasible in practice.

Our goal is to further investigate the relationship between calibration and error rates.
We show that even if the Equalized Odds conditions are relaxed
substantially -- requiring only that weighted sums of the group error
rates match -- it is still problematic to also enforce calibration.
We provide necessary and sufficient conditions under which this calibrated
relaxation
is feasible.
When feasible, it has a unique optimal solution that
can be achieved through post-processing of existing classifiers.
Moreover, we provide a simple post-processing algorithm to find this solution:
withholding predictive information for randomly chosen inputs
to achieve parity and preserve calibration. However, this simple
post-processing method is fundamentally unsatisfactory: although the
post-processed predictions of our information-withholding algorithm are ``fair'' in
expectation, most practitioners would object to the fact that a non-trivial
portion of the individual predictions are withheld as a result of coin tosses -- especially in
sensitive settings such as health care or criminal justice. The
optimality of this algorithm thus has troubling implications and shows that
calibration and error-rate fairness are inherently at odds (even beyond the initial results
by \cite{chouldechova2017fair} and \cite{kleinberg2016inherent}).

Finally, we evaluate these theoretical findings empirically,
comparing calibrated notions of non-discrimination against
the (uncalibrated) Equalized Odds framework on several datasets.
These experiments further support our conclusion that calibration and error-rate constraints are in most cases mutually incompatible goals.
In practical settings, it may be advisable to choose only one of these goals rather than attempting to achieve some relaxed notion of both.

\section{Related Work}

\paragraph{Calibrated probability estimates} are considered necessary for empirical risk
analysis tools \cite{berk2017fairness,crowson-calibration-risk-scores,dieterich-northpointe-fairness,flores-re-propublica-fair}.
In practical applications, uncalibrated probability estimates can be misleading
in the sense that the end user of these estimates has an incentive to mistrust (and therefore potentially misuse) them.
We note however that calibration does not remove all potential for misuse, as the end user's biases
might cause her or him to treat estimates differently based on group membership.
There are several post-processing methods for producing calibrated outputs from classification
algorithms. For example, Platt Scaling \cite{platt1999probabilistic} passes outputs through
a learned sigmoid function, transforming them into calibrated probabilities.
Histogram Binning and Isotonic Regression \cite{zadrozny2001obtaining}
learn a general monotonic function from outputs to probabilities.
See \cite{niculescu2005predicting} and \cite{guo2017calibration} for empirical
comparisons
of these methods.

\paragraph{Equalized Odds}~\cite{hardt2016equality},
also referred to as \emph{Disparate Mistreatment} \cite{zafar2017fairness},
ensures that no error type disproportionately affects any particular group.
\citet{hardt2016equality} provide a post-processing technique to achieve this
framework, while \citet{zafar2017fairness} introduce optimization constraints
to achieve non-discrimination at training time.
Recently, this framework has received significant attention from the algorithmic
fairness community.
Researchers have found that it is incompatible with other notions
of fairness \cite{kleinberg2016inherent,chouldechova2017fair,corbett2017algorithmic}.
Additionally, \citet{woodworth2017learning} demonstrate that, under certain assumptions, post-processing
methods for achieving non-discrimination may be suboptimal.

\paragraph{Alternative fairness frameworks} exist and are continuously proposed.
We highlight several of these works, though by no means offer
a comprehensive list. (More thorough reviews can be found in
\cite{barocas2016big,berk2017fairness,romei2014multidisciplinary}).
It has been shown that, under most frameworks of fairness, there is a trade-off
between algorithmic performance and non-discrimination
\cite{berk2017fairness,corbett2017algorithmic,hardt2016equality,zliobaite2015relation}.
Several works approach fairness through the lens of
\emph{Statistical Parity}
\cite{calders2009building,kamiran2009classifying,calders2010three,kamishima2011fairness,zemel2013learning,louizos2015variational,edwards2015censoring,johndrow2017algorithm}.
Under this definition, group membership should not affect the prediction of a classifier,
i.e. members of different groups should have the same probability of receiving
a positive-class prediction.
%
However,
it has been argued that Statistical Parity may not be applicable
in many scenarios
\cite{chouldechova2017fair,dwork2012fairness,hardt2016equality,kleinberg2016inherent},
as it
%
attempts to guarantee equal representation. For example, it is
inappropriate in criminal justice, where base rates differ
across different groups.
A related notion is \emph{Disparate Impact}
\cite{feldman2015certifying,zafar2015learning}, which states that the
prediction rates for any two groups should not differ by more than $80\%$
(a number motivated by legal requirements).
\citet{dwork2012fairness} introduce a notion of fairness based on
the idea that similar individuals should receive similar outcomes,
though it challenging to achieve this notion in practice.
Fairness has also been considered in online learning
\cite{joseph2016fairness,kearns2017meritocratic}, unsupervised learning \cite{bolukbasi2016man},
and causal inference \cite{kilbertus2017avoiding,kusner2017counterfactual}.


\section{Problem Setup}
\label{sec:setup}

The setup of our framework most follows the \emph{Equalized Odds}
framework \cite{hardt2016equality,zafar2017fairness}; however, we extend their framework for use with
probabilistic classifiers.
Let $P \subset \R^k \times \{0, 1\}$ be the input space of a binary classification task.
In our criminal justice example, $\left( \x, y \right) \sim P$ represents a person, with
$\x$ representing the individual's history and $y$ representing whether or not the person
will commit another crime.
Additionally, we assume the presence of two groups
$G_1, G_2 \subset P$,
which represent disjoint population subsets, such as different races.
We assume that the groups have different \emph{base rates} $\br_{t}$, or probabilities
of belonging to the positive class:
$
\br_{1} = \Pr_{(\x, y) \sim G_1} \left[ y = 1 \right] \neq \Pr_{(\x, y) \sim G_2} \left[ y = 1 \right] = \br_{2}.
$

Finally, let $h_1, h_2 : \R^k \rightarrow [0, 1] $ be binary classifiers,
where $h_1$ classifies samples from $G_1$ and $h_2$ classifies samples from $G_2$.\footnote{
In practice, $h_1$ and $h_2$ can be trained jointly (i.e. they are the same classifier).}
Each classifier outputs the probability that a given sample $\x$ belongs to the positive class.
The notion of Equalized Odds non-discrimination is based on the false-positive
and false-negative rates for each group, which we generalize here for use with probabilistic
classifiers:
\begin{definition}
  \label{def:fpfn}
  The \emph{generalized false-positive rate} of classifier $h_t$ for group $G_t$ is
  $
  \fp{h_t} = \ev_{(\x, y) \sim G_t} \bigl[ h_t(\x) \mid y\!=\!0 \bigr].
  $
  Similarly, the \emph{generalized false-negative rate} of classifier $h_t$ is
  $
    \fn{h_t} = \ev_{(\x, y) \sim G_t} \bigl[ \left( 1 - h_t(\x) \right) \mid y\!=\!1 \bigr].
  $
\end{definition}
If the classifier were to output either $0$ or $1$, this represents
the standard notions of false-positive and false-negative rates.
We now define the Equalized Odds framework (generalized for probabilistic classifiers),
which aims to ensure that errors of a given type are not biased against
any group.
\begin{definition}[Probabilistic Equalized Odds]
  \label{def:eo}
  Classifiers $h_1$ and $h_2$ exhibit Equalized Odds for groups $G_1$
  and $G_2$ if $\fp{h_1} = \fp{h_2}$ and $\fn{h_1} = \fn{h_2}$.
\end{definition}

\paragraph{Calibration Constraints.}
As stated in the introduction, these two conditions do not necessarily prevent
discrimination if the classifier predictions do not represent well-calibrated
probabilities.
Recall that calibration intuitively says that
probabilities should carry semantic meaning:
if there are 100 people in $G_1$ for whom
$h_1(\x) = 0.6$, then we expect $60$ of them to belong to the positive class.
\begin{definition}
  \label{def:calibration}
  A classifier $h_t$ is \emph{perfectly calibrated} if $\forall p \in [0, 1]$,
  $\Pr_{(\x, y) \sim G_t} \bigl[ y\!=\!1 \mid h_t(\x)\!=\!p \bigr] = p$.
\end{definition}
It is commonly accepted amongst practitioners that both classifiers $h_1$ and $h_2$
should be calibrated \emph{with respect to groups} $G_1$ and $G_2$ to prevent
discrimination \cite{berk2017fairness,crowson-calibration-risk-scores,dieterich-northpointe-fairness,flores-re-propublica-fair}.
Intuitively, this prevents the probability scores from carrying group-specific
information.
Unfortunately, \citet{kleinberg2016inherent} (as well as
\cite{chouldechova2017fair}, in a binary setting) prove that
a classifier cannot achieve both calibration and Equalized Odds, even in an approximate
sense, except in the most
trivial of cases.

\subsection{Geometric Characterization of Constraints}

\begin{figure}[t]
  \centering
  \begin{subfigure}[b]{0.22\textwidth}
    \includegraphics[width=\textwidth]{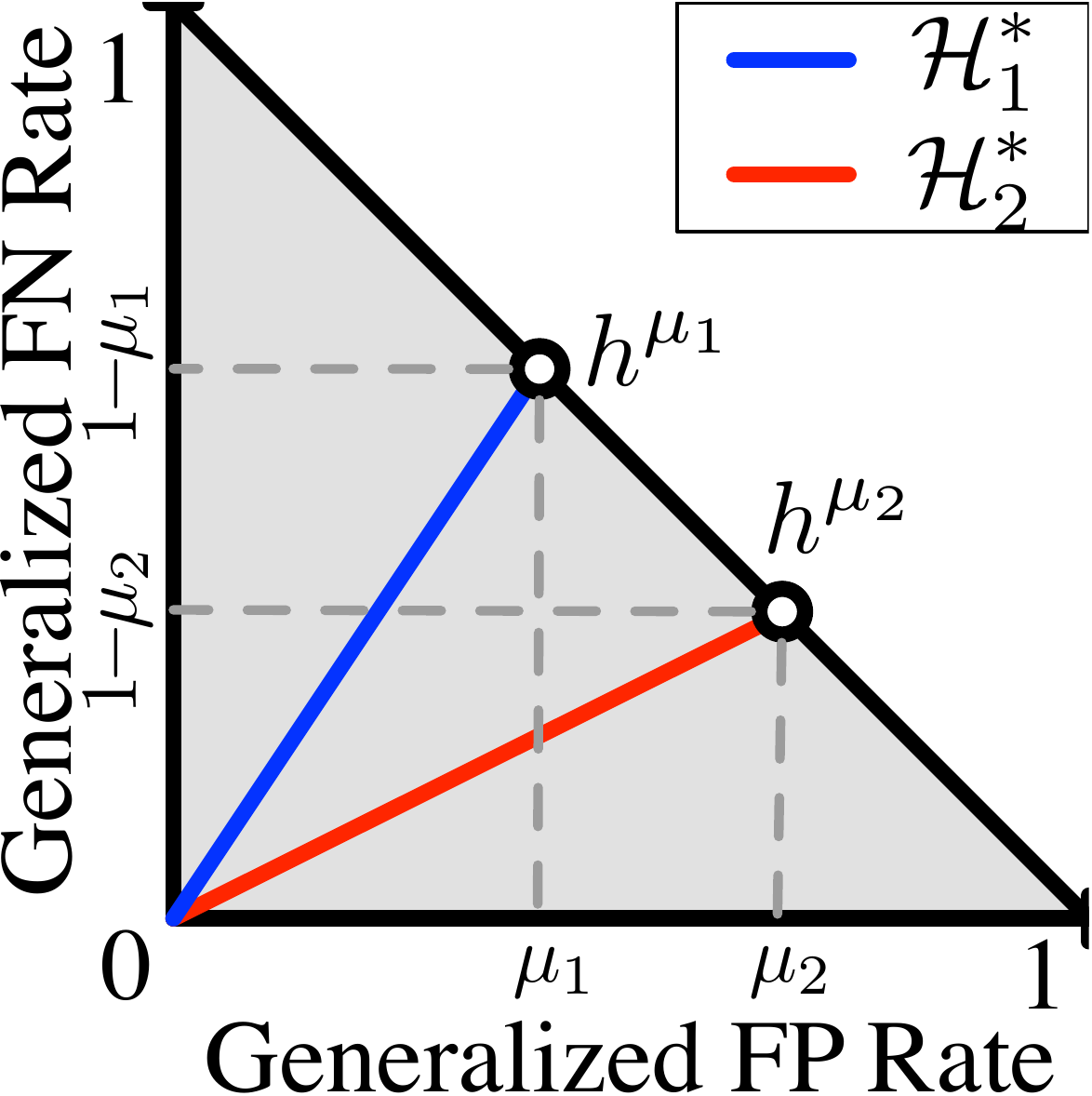}
    \caption{Possible cal. classifiers $\mathcal{H}^*_1, \mathcal{H}^*_2$ (blue/red).}
    \label{fig:calibration}
  \end{subfigure}
  \quad
  \begin{subfigure}[b]{0.22\textwidth}
    \includegraphics[width=\textwidth]{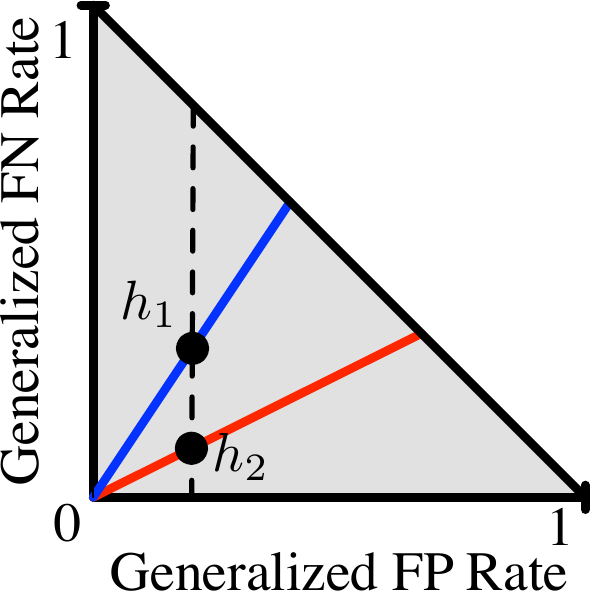}
    \caption{Satisfying cal. and equal F.P. rates.}
    \label{fig:two-of-three-fp}
  \end{subfigure}
  \quad
  \begin{subfigure}[b]{0.22\textwidth}
    \includegraphics[width=\textwidth]{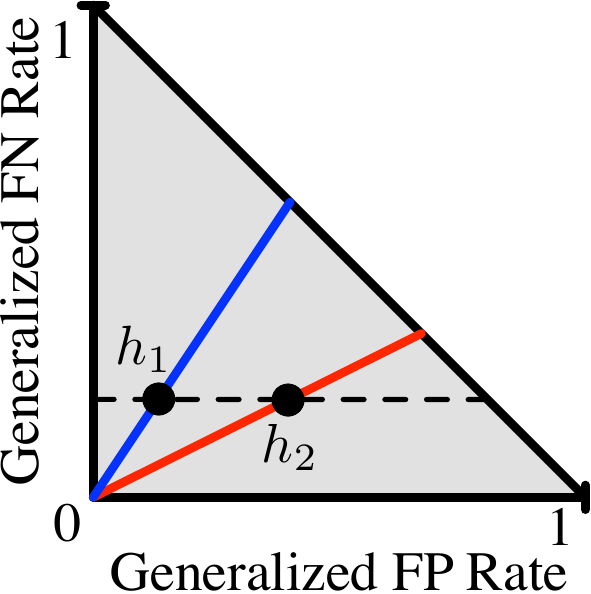}
    \caption{Satisfying cal. and equal F.N. rates.}
    \label{fig:two-of-three-fn}
  \end{subfigure}
  \quad
  \begin{subfigure}[b]{0.22\textwidth}
    \includegraphics[width=\textwidth]{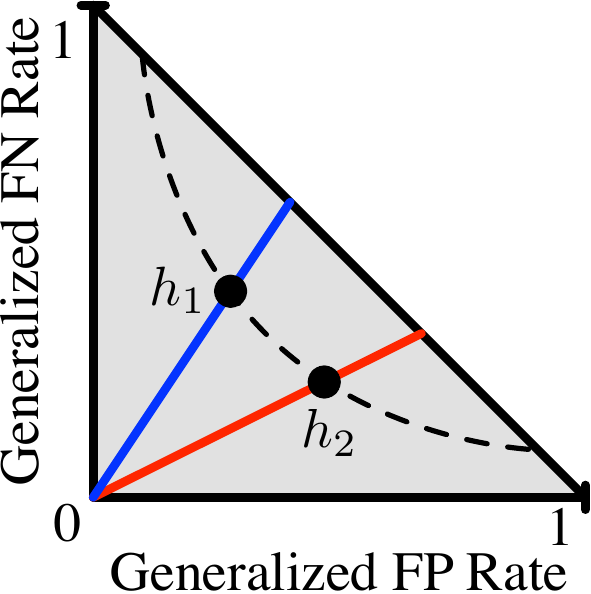}
    \caption{Satisfying cal. and a general constraint.}
    \label{fig:general-disparity}
  \end{subfigure}
  \caption{
  Calibration, trivial classifiers, and equal-cost constraints -- plotted in the
  false-pos./false-neg. plane.
  $\mathcal{H}^*_1, \mathcal{H}^*_2$ are the set of cal. classifiers for the two groups, and $h^{\mu_1}, h^{\mu_2}$ are trivial classifiers.
  }
  \label{fig:two-of-three-conditions}
\end{figure}

We now will characterize the calibration and error-rate constraints with
simple geometric intuitions. Throughout the rest of this paper, all of our
results can be easily derived from this interpretation.
We begin by defining the region of classifiers which are \emph{trivial}, or
those that output a constant value for all inputs (i.e. $h^{c}(\x) = c$,
where $0 \leq c \leq 1$ is a constant).
We can visualize these classifiers on a graph with generalized false-positive
rates on one axis and generalized false-negatives on the other.
It follows from the definitions of generalized false-positive/false-negative
rates and calibration that all trivial classifiers $h$ lie on the diagonal defined by
$\fp{h} + \fn{h} = 1$ (\autoref{fig:calibration}).
Therefore, all classifiers that are ``better than random'' must lie below this
diagonal in false-positive/false-negative space (the gray triangle in the figure).
Any classifier that lies above the diagonal performs
``worse than random,'' as we can find a point on the trivial classifier diagonal
with lower false-positive and false-negative rates.

Now we will characterize the set of calibrated classifiers for groups $G_1$ and $G_2$, which
we denote as $\mathcal{H}^*_1$ and $\mathcal{H}^*_2$.
\citeauthor{kleinberg2016inherent} show that the generalized false-positive and
false-negative rates of a calibrated classifier are
linearly related by the base rate of the group:\footnote{
  Throughout this work we will treat the calibration constraint as holding
  exactly; however, our results generalize to approximate settings as well. See the
  Supplementary Materials for more details.
}
\begin{equation}
  \fn{h_t} = (1 - \br_t) / \br_t \: \fp{h_t}.
\end{equation}
In other words, $h_1$ lies on
a line with slope $(1 - \br_1) / \br_1$ and
$h_2$ lies on a line with slope $(1 - \br_2) / \br_2$ (\autoref{fig:calibration}).
The lower endpoint of each line is the \emph{perfect classifier}, which
assigns the correct prediction with complete certainty to every input.
The upper endpoint is a trivial classifier, as no calibrated classifier
can perform ``worse than random'' (see \autoref{lma:trivial} in
\autoref{app:cost-funcs}).
The only trivial classifier that satisfies the calibration
condition for a group $G_t$ is the one that outputs the base rate $\mu_t$.
We will refer to $h^{\br_1}$ and $h^{\br_2}$ as the trivial classifiers,
calibrated for groups $G_1$ and $G_2$ respectively.
It follows from the definitions that $\fp{h^{\br_1}} = \mu_1$
and $\fn{h^{\br_1}} = 1 - \mu_1$, and likewise for $h^{\br_2}$.

Finally, it is worth noting that for calibrated classifiers,
a lower false-positive rate necessarily corresponds to a lower false-negative rate
and vice-versa.
In other words, for a given base rate, a ``better'' calibrated classifier
lies closer to the origin on the line of calibrated classifiers.

\paragraph{Impossibility of Equalized Odds with Calibration.}
With this geometric intuition, we can provide a simplified
proof of the main impossibility
result from \cite{kleinberg2016inherent}:
\begin{thm*}[Impossibility Result \cite{kleinberg2016inherent}]
  \label{thm:orig-impossibility}
  Let $h_1$ and $h_2$ be classifiers for groups $G_1$ and $G_2$ with $\br_1 \neq \br_2$.
  $h_1$ and $h_2$ satisfy the Equalized Odds and calibration conditions
  if and only if $h_1$ and $h_2$ are perfect predictors.
\end{thm*}
Intuitively, the three conditions define a set of classifiers which
is overconstrained.
Equalized Odds stipulates that the classifiers $h_1$ and
$h_2$ must lie on the same coordinate in the false-positive/false-negative
plane. As $h_1$ must lie on the blue line of calibrated classifiers for $\mathcal{H}_1^*$ and $h_2$ on the red line $\mathcal{H}_2^*$ they can only satisfy EO at the unique intersection point --- the origin (and location of the perfect classifier).
This implies that unless the two classifiers achieve perfect accuracy,
we must relax the Equalized Odds conditions if we want to
maintain calibration.

\begin{figure}[t]
  \centering
  \begin{subfigure}[b]{0.22\textwidth}
    \includegraphics[width=\textwidth]{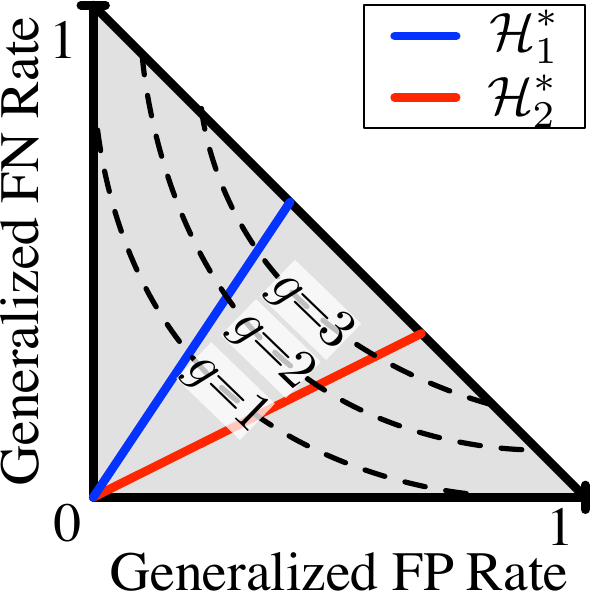}
    \caption{Level-order curves of cost. Low cost implies low
      error rates.}
    \label{fig:levelorder}
  \end{subfigure}
  \quad
  \begin{subfigure}[b]{0.22\textwidth}
    \includegraphics[width=\textwidth]{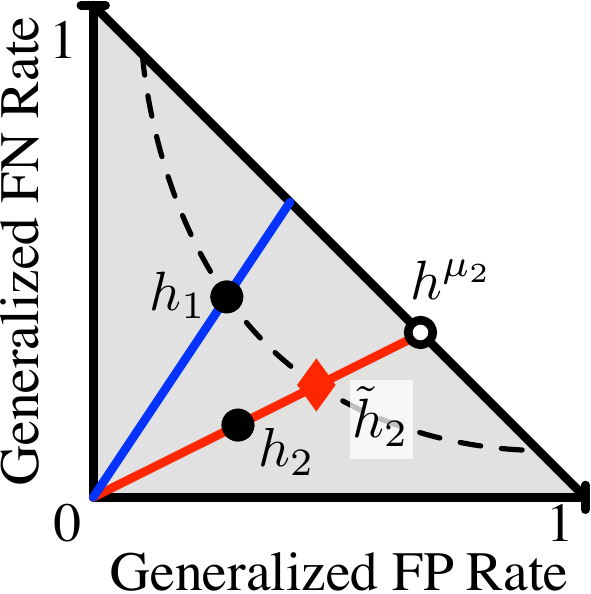}
    \caption{Usually, there is a calibrated classifier $\tilde h_2$ with the same cost of $h_1$.}
    \label{fig:postprocess}
  \end{subfigure}
  \quad
  \begin{subfigure}[b]{0.22\textwidth}
    \includegraphics[width=\textwidth]{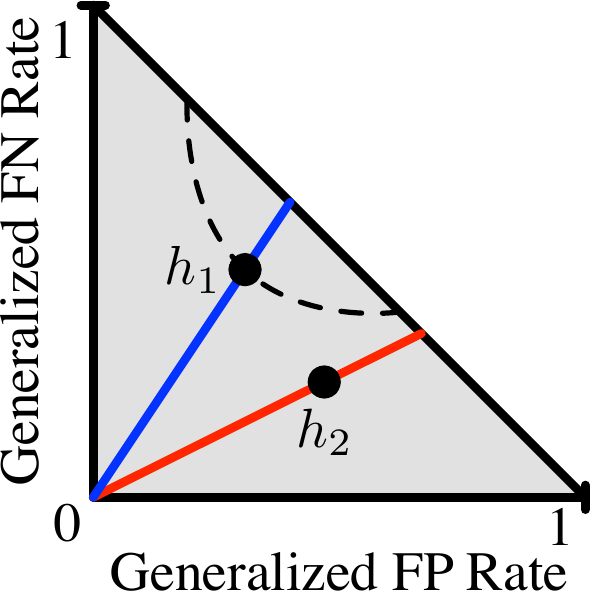}
    \caption{Cal. and equal-cost are incompatible
    if $h_1$ has high error.}
    \label{fig:infeasible-trivial}
  \end{subfigure}
  \quad
  \begin{subfigure}[b]{0.22\textwidth}
    \includegraphics[width=\textwidth]{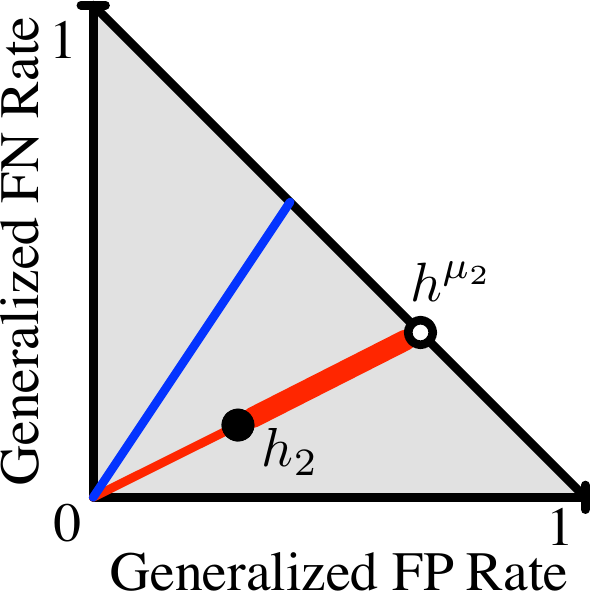}
    \caption{Possible cal. classifiers for $G_2$ (bold red) by
    mixing $h_2$ and $h^{\mu_2}$.}
    \label{fig:mixing}
  \end{subfigure}
  \caption{
    Calibration-Preserving Parity through interpolation.
  }
  \label{fig:feasible}
\end{figure}


\section{Relaxing Equalized Odds to Preserve Calibration}
\label{sec:relaxation}

In this section, we show that a substantially simplified notion of Equalized Odds
is compatible with calibration.
We introduce a general relaxation that seeks
to satisfy a \emph{single equal-cost constraint} while maintaining calibration for each group $G_t$.
%
We begin with the observation that Equalized Odds
sets constraints to equalize
false-positives $\fp{h_t}$ and false-negatives $\fn{h_t}$. To capture and
generalize this, we define a
\emph{cost function} \fairconstname{} to be a linear function in \fp{h_t} and
\fn{h_t} with arbitrary dependence on the group's base rate $\br_t$.
More formally, a cost function for group $G_t$ is
\begin{equation}
  \fairconst{h_t}{t} = a_t \fp{h_t} + b_t \fn{h_t}
  \label{eqn:cost_general}
\end{equation}
where $a_{t}$ and $b_{t}$ are non-negative constants that are specific to each group (and
thus may depend on $\br_t$): see \autoref{fig:general-disparity}. We also make
the assumption that for any $\br_t$,
at least one of $a_t$ and $b_t$ is nonzero, meaning
$\fairconst{h_t}{t} = 0$ if and only if $\fp{h_t} = \fn{h_t} = 0$.\footnote{
  By calibration, we cannot have one of $\fp{h_t} = 0$ or $\fn{h_t} = 0$
  without the other, see \autoref{fig:calibration}.
}
This class of cost functions encompasses a
variety of scenarios.
As an example, imagine an application in which the equal false-positive
condition is essential but not the
false-negative condition.
Such a scenario may arise in our recidivism-prediction example, if we require
that non-repeat offenders of any race are not disproportionately labeled as high risk.
If we plot the set of calibrated classifiers $\mathcal{H}^*_1$ and $\mathcal{H}^*_2$
on the false-positive/false-negative plane, we can see that ensuring the
false-positive condition requires finding classifiers $h_1 \in \mathcal{H}^*_1$
and $h_2 \in \mathcal{H}^*_2$ that fall on the same vertical line
(\autoref{fig:two-of-three-fp}). Conversely, if we instead choose to satisfy only
the false-negative condition, we would find classifiers $h_1$ and $h_2$ that
fall on the same horizontal (\autoref{fig:two-of-three-fn}).
Finally, if both false-positive and false-negative errors
incur a negative cost on the individual, we may choose to equalize a weighted
combination of the error rates \cite{berk2016primer,berk2017fairness,chouldechova2017fair},
which can be graphically described by the classifiers lying on a convex and negatively-sloped level set
(\autoref{fig:general-disparity}).
With these definitions,
we can formally define our relaxation:
\begin{definition}[Relaxed Equalized Odds with Calibration]
  \label{def:fair}
  Given a cost function $\fairconstname$ of the form in $\eqref{eqn:cost_general}$,
  classifiers $h_1$ and $h_2$ achieve \emph{Relaxed Equalized Odds with
  Calibration} for groups $G_1$ and $G_2$ if both classifiers
  are calibrated and satisfy the constraint $\fairconst{h_1}{1} =
  \fairconst{h_2}{2}$.
\end{definition}
It is worth noting that, for calibrated classifiers, an increase in cost
strictly corresponds to an increase in both the false-negative and
false-positive rate. This can be interpreted graphically, as the level-order cost
curves lie further away from the origin as cost increases (\autoref{fig:levelorder}). In other words,
the cost function can always be used as a proxy for either error rate.\footnote{
  This holds even for approximately calibrated classifiers ---  see \autoref{sec:cost_and_error}.
}

\paragraph{Feasibility.}
It is easy to see that \autoref{def:fair} is always satisfiable -- in
Figures~\ref{fig:two-of-three-fp}, \ref{fig:two-of-three-fn}, and \ref{fig:general-disparity}
we see that there are many such solutions that would lie on a given level-order
cost curve while maintaining calibration, including the case in which both
classifiers are perfect.  In
practice, however, not all classifiers are achievable. For the rest of
the paper, we will assume that we have access to ``optimal'' (but possibly
discriminatory) calibrated
classifiers $h_1$ and $h_2$ such that, due to whatever limitations there are on
the predictability of the task, we are unable to find other classifiers that
have lower cost with respect to $\fairconstname$.
We allow $h_1$ and
$h_2$ to be learned in any way, as long as they are calibrated. Without loss of
generality, for the remainder of the paper, we will assume that
$\fairconst{h_1}{1} \ge \fairconst{h_2}{2}$.

Since by assumption we have no way to find a classifier for $G_1$ with lower
cost than $h_1$,
our goal is therefore to find
a classifier $\fair h_2$ with cost equal to $h_1$. This pair of classifiers
would represent the lowest cost (and therefore optimal) set of classifiers that
satisfies calibration and the equal cost constraint.
For a given base rate $\mu_t$ and value of the cost
function $\fairconstname$, a calibrated classifier's position in the
generalized false-positive/false-negative plane is uniquely determined
(\autoref{fig:levelorder}).
This is because each level-order
curve of the cost function $\fairconstname$ has negative
slope in this plane, and each level order curve only intersects a group's calibrated
classifier line once.
In other words, there is a unique solution in the false-positive/false-negative
plane for classifier $\fair h_2$ (\autoref{fig:postprocess}).

Consider the range of values that $\fairconstname$ can take. As noted above,
$\fairconst{h_t}{t} \ge 0$, with equality if and only if $h_t$ is the perfect
classifier. On the other hand,
the trivial classifier
(again, which outputs the constant $\br_t$ for all inputs)
is the calibrated classifier that achieves maximum cost for any $\fairconstname$
(see \autoref{lma:trivial} in \autoref{app:cost-funcs}).
As a result, the cost of a classifier for group
$G_t$ is between 0 and $\fairconst{h^{\br_t}}{t}$.
This naturally leads to a characterization of feasibility: \autoref{def:fair}
can be achieved if and only if $h_1$ incurs less cost than group $G_2$'s
trivial classifier $h^{\br_2}$; i.e.
if $\fairconst{h_1}{1} \leq \fairconst{h^{\mu_2}}{2}$.
This can be seen graphically in Figure \ref{fig:infeasible-trivial},
in which the level-order curve for $\fairconst{h_1}{1}$ does not intersect the set
of calibrated classifiers for $G_2$. Since, by assumption, we cannot find a
calibrated classifier for $G_1$ with
strictly smaller cost than $h_1$, there is no feasible solution.
On the other hand, if $h_1$ incurs less cost than $h^{\br_2}$, then we
will show feasibility by construction with a simple algorithm.

\paragraph{An Algorithm.}
While it may be possible to encode the constraints of \autoref{def:fair} into the training
procedure of $h_1$ and $h_2$, it is not immediately obvious how to do so.
Even naturally probabilistic algorithms, such as logistic regression,
can become uncalibrated in the presence of optimization constraints (as
is the case in \cite{zafar2017fairness}). It is not straightforward
to encode the calibration constraint if the probabilities
are assumed to be continuous, and post-processing calibration methods
\cite{platt1999probabilistic,zadrozny2001obtaining}
would break equal-cost constraints by modifying classifier scores.
Therefore, we look to achieve the calibrated Equalized Odds relaxation
by post-processing existing calibrated classifiers.

Again, given $h_1$ and $h_2$ with $\fairconst{h_1}{1} \ge \fairconst{h_2}{2}$, we want
to arrive at a calibrated classifier $\fair h_2$ for group $G_2$ such that
$\fairconst{h_1}{1} = \fairconst{\fair h_2}{2}$. Recall that, under our
assumptions, this would be
the best possible solution with respect to classifier cost. We show that this cost
constraint can be achieved by withholding predictive information for a randomly
chosen subset of group $G_2$. In other words, rather than always returning $h_2(\x)$
for all samples, we will occasionally return the group's mean probability
(i.e. the output of the trivial classifier $h^{\mu_2}$). In
\autoref{lma:interpolation} in \autoref{app:cost-funcs}, we show
that if
\begin{equation}
  \label{eqn:interpolation}
  \fair h_2(\x) = \begin{cases}
    h^{\br_2}(\x) = \br_2 & \text{with probability } \alpha \\
    h_2(\x) & \text{with probability } 1-\alpha
  \end{cases}
\end{equation}
then the cost of $\fair h_2$ is a linear interpolation between the costs of
$h_2$ and $h^{\br_2}$ (\autoref{fig:mixing}). More formally, we
have that $\fairconst{\fair h_2}{2} = (1-\alpha)
\fairconst{h_2}{2} + \alpha \fairconst{h^{\br_2}}{2}$), and thus setting $\alpha =
\frac{\fairconst{h_1}{1} - \fairconst{h_2}{2}}{\fairconst{h^{\br_2}}{2} -
\fairconst{h_2}{2}}$ ensures that $\fairconst{\fair h_2}{2} =
\fairconst{h_1}{1}$ as desired (\autoref{fig:postprocess}).
Moreover, this randomization preserves calibration (see
\autoref{sec:optimality-proofs}).
\autoref{alg:method} summarizes this method.
\begin{algorithm}
  \caption{Achieving Calibration and an Equal-Cost Constraint via Information Withholding}
  \label{alg:method}
  \begin{algorithmic}
    \STATE {\bfseries Input:} classifiers $h_1$ and $h_2$ s.t.
    $\fairconst{h_2}{2} \leq \fairconst{h_1}{1} \leq \fairconst{h^{\br_2}}{2}$, holdout set $P_{valid}$.

    \begin{itemize}
      \STATE Determine base rate $\br_2$ of $G_2$ (using
        $P_{valid}$) to produce trivial classifier $h^{\br_2}$.

      \STATE Construct $\fair h_2$ using with
        $ \alpha = \frac{\fairconst{h_1}{1} - \fairconst{h_2}{2}}{\fairconst{h^{\br_2}}{2} - \fairconst{h_2}{2}}$,
        where $\alpha$ is the interpolation parameter.
    \end{itemize}

    \RETURN $h_1$, $\fair h_2$ --- which are calibrated and satisfy $\fairconst{h_1}{1} = \fairconst{\fair h_2}{2}$.
  \end{algorithmic}
\end{algorithm}


\paragraph{Implications.}
In a certain sense, \autoref{alg:method} is an ``optimal'' method
because it arrives at the unique false-negative/false-positive solution for $\fair h_2$,
where $\fair h_2$ is calibrated and has cost equal to $h_1$. Therefore
(by our assumptions) we can find no better classifiers that satisfy \autoref{def:fair}.
%
This simple result has strong consequences, as the tradeoffs to
satisfy both calibration and the equal-cost constraint are often unsatisfactory
--- both intuitively and experimentally (as we will show in \autoref{sec:experiments}).

We find two primary objections to this solution. First, it equalizes
costs simply by making a classifier strictly worse for one of the groups.
Second, it achieves this
cost increase by withholding information on a randomly chosen population subset,
making the outcome inequitable within the group (as measured by a standard measure
of inequality like the Gini coefficient). Due to the optimality of the
algorithm, the former of these issues is unavoidable in \emph{any} solution that
satisfies \autoref{def:fair}. The latter, however, is slightly more subtle, and
brings up the question of \emph{individual fairness} (what guarantees we would
like an algorithm to make with respect to each individual) and how it interacts
with \emph{group fairness} (population-level guarantees). While this certainly
is an important issue for future work, in this particular setting, even if one
could find another algorithm that distributes the burden of additional cost more
equitably, any algorithm will make at least as many false-positive/false-negative
errors as \autoref{alg:method}, and these misclassifications will always be
tragic to the individuals whom they affect.
The performance loss across the entire group is often
significant enough to make this combination of constraints somewhat worrying to
use in practice, regardless of the algorithm.

\paragraph{Impossibility of Satisfying Multiple Equal-Cost Constraints.}
\label{ssec:impossibility}
It is natural to argue there might be multiple cost functions that we
would like to equalize across groups.
However, satisfying more than one distinct
equal-cost constraint (i.e. different curves in the F.P./F.N. plane) is infeasible.
%
\begin{thm}[Generalized impossibility result] \label{thm:impossibility}
  Let $h_1$ and $h_2$ be calibrated classifiers for $G_1$ and $G_2$ with equal
  cost with respect to
  $\fairconstname$.
  If $\mu_1 \ne \mu_2$, and if $h_1$ and $h_2$ also have equal cost with
  respect to a different cost function $\fairconstname'$, then $h_1$ and $h_2$
  must be perfect classifiers.
\end{thm}
(Proof in \autoref{app:impossiblility}). Note that this is
a generalization of the impossibility result of \cite{kleinberg2016inherent}.
Furthermore, we show in \autoref{thm:approx-impossibility} (in \autoref{app:impossiblility})
that this holds in an
approximate sense: if calibration and multiple distinct equal-cost constraints
are approximately achieved by some classifier, then that classifier must have
approximately zero generalized false-positive and false-negative rates.

%

\section{Experiments}
\label{sec:experiments}

\begin{figure}[t!]
  \centering
  \begin{subfigure}[b]{0.46\textwidth}
    \caption{Income Prediction.}
    \includegraphics[width=\columnwidth]{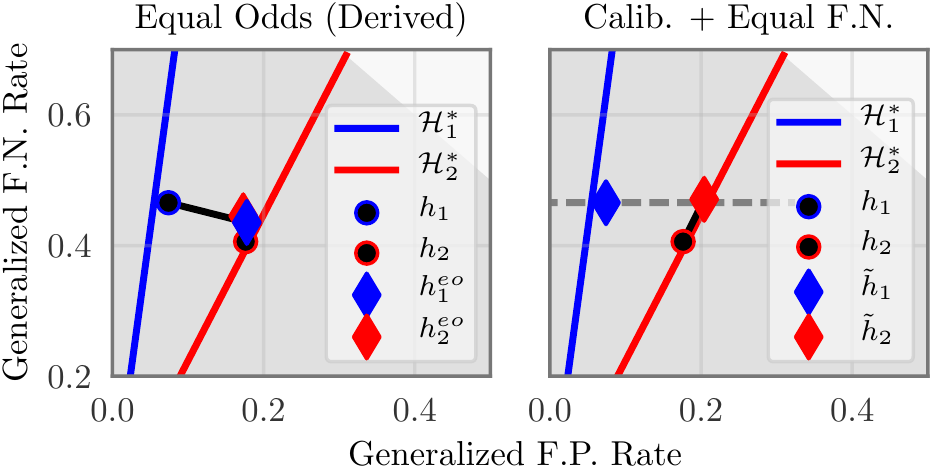}
    \label{fig:adult-res}
  \end{subfigure}
  \quad
  \begin{subfigure}[b]{0.46\textwidth}
    \caption{Health Prediction.}
    \includegraphics[width=\columnwidth]{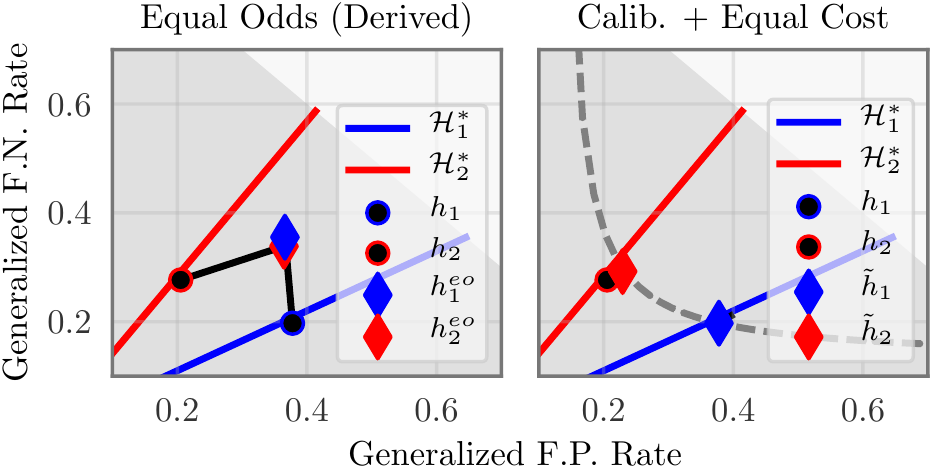}
    \label{fig:health-res}
  \end{subfigure}
  \begin{subfigure}[b]{0.67\textwidth}
    \caption{Recidivism Prediction.}
    \includegraphics[width=\columnwidth]{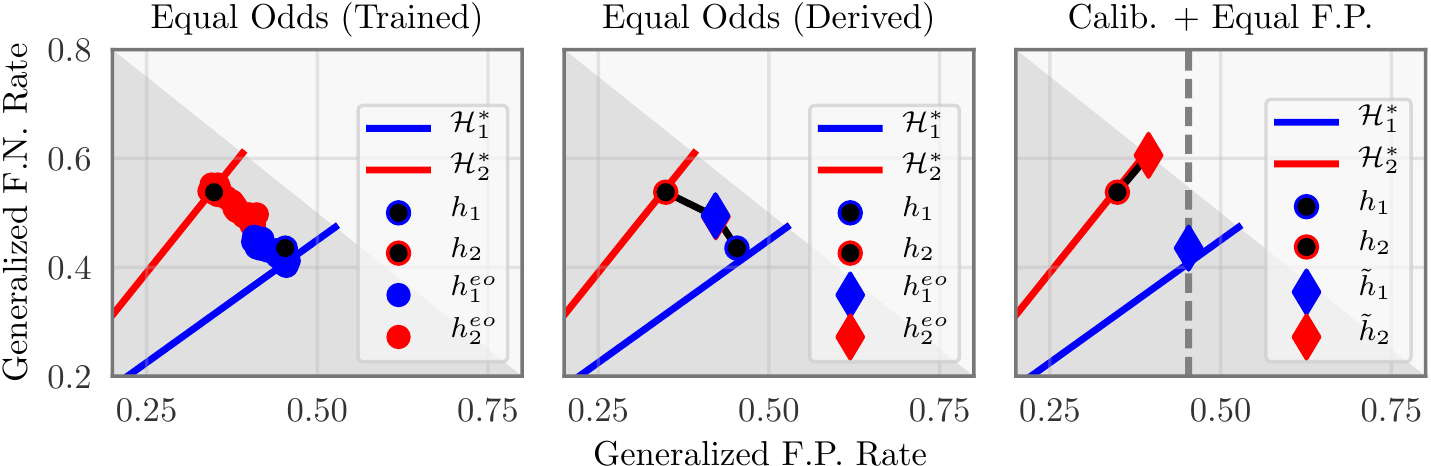}
    \label{fig:compas-res}
  \end{subfigure}
  \vspace{-1.3em}
  \caption{Generalized F.P. and F.N. rates for two groups under
  Equalized Odds and the calibrated relaxation. Diamonds represent post-processed classifiers.
  Points on the Equalized Odds (trained) graph represent classifiers achieved by modifying constraint hyperparameters.}
  \label{fig:res}
\end{figure}


In light of these findings, our goal is to understand the impact
of imposing calibration and an equal-cost constraint on
real-world datasets. We will empirically show that, in many cases, this will
result in performance degradation, while simultaneously increasing other
notions of disparity.
We perform experiments on three datasets: an
income-prediction, a health-prediction, and a criminal
recidivism dataset.  For each task, we
choose a cost function within our framework
that is appropriate for the given scenario.
We begin with two calibrated classifiers $h_1$ and $h_2$
for groups $G_1$ and $G_2$.
We assume that these classifiers cannot be significantly
improved without more training data or features.
We then derive $\fair h_2$ to equalize the costs while maintaining calibration.
The original classifiers are trained on a portion of the data, and then the new classifiers are
derived using a separate holdout set.
To compare against the (uncalibrated) Equalized Odds framework, we derive
F.P./F.N. matching classifiers using the post-processing
method of \cite{hardt2016equality} ({\bf \hardtetal{}}).
On the criminal recidivism dataset, we additionally learn classifiers that directly
encode the Equalized Odds constraints, using the methods of
\cite{zafar2017fairness} ({\bf \zafaretal{}}).
(See \autoref{app:experiments} for detailed training and post-processing procedures.)
We visualize model error rates on the
generalized F.P. and F.N. plane.
Additionally, we plot the calibrated
classifier lines for $G_1$ and $G_2$ to visualize model calibration.

\paragraph{Income Prediction.}
  The Adult Dataset from UCI Machine Learning Repository \citep{Lichman:2013}
  contains 14 demographic and occupational features for various people, with the
  goal of predicting whether a person's income is above $\$50,000$.
  In this scenario, we seek to achieve predictions with equalized cost
  across genders ($G_1$ represents women and $G_2$ represents men).
  We model a scenario where the primary concern is ensuring
  equal generalized F.N. rates across genders,
  which would, for example, help job recruiters  prevent gender discrimination
  in the form of
  underestimated salaries.
  Thus, we choose our cost constraint to require equal generalized
  F.N. rates across groups.
  In \autoref{fig:adult-res}, we see that the original classifiers $h_1$ and
  $h_2$ approximately
  lie on the line of calibrated classifiers.
  In the left plot (\hardtetal{}), we see that it is possible to
  (approximately) match both error rates of the classifiers at the cost of $\eo
  h_1$ deviating from the set of calibrated classifiers.
  In the right plot,
  we see that it is feasible to
  equalize the generalized F.N. rates while maintaining calibration.
  $h_1$ and $\fair h_2$ lie on the same
  level-order curve of $\fairconstname$ (represented by the dashed-gray line),
  and simultaneously remain on the ``line'' of calibrated classifiers.
  It is worth noting that achieving either notion of non-discrimination requires
  some cost to at least one of the groups.
  However, maintaining calibration further increases the difference
  in F.P. rates between groups. In some sense, the calibrated framework
  trades off one notion of disparity for another while simultaneously increasing the overall error rates.

\paragraph{Health Prediction.}
  The Heart Dataset from the UCI Machine Learning Repository contains 14 processed features
  from 906 adults in 4 geographical locations. The goal of this dataset is to
  accurately predict whether or not an
  individual has a heart condition. In this scenario, we would like to reduce
  disparity between
  middle-aged adults ($G_1$) and seniors ($G_2$).
  In this scenario, we consider F.P. and F.N. to both be undesirable.
  A false prediction of a heart condition could result in unnecessary medical attention, while
  false negatives incur cost from delayed treatment. We therefore utilize the following cost function
  $
    \fairconst{h_t}{t} = r_{fp} h_t(\x) \left(1 - y\right) + r_{fn} \left( 1 - h_t(\x) \right) y,
  $
  which essentially assigns a weight to both F.N. and F.P. predictions.
  In our experiments, we set $r_{fp}=1$ and $r_{fn}=3$.
  In the right plot of \autoref{fig:health-res}, we can see that
  the level-order curves of the cost function form a curved
  line in the generalized F.P./F.N. plane.
  Because our
  original classifiers lie approximately
  on the same level-order curve, little change is required to equalize the costs of
  $h_1$ and $\fair h_2$ while maintaining calibration.
  This is the only experiment in which the calibrated framework
  incurs little additional cost, and therefore could be considered a viable option.
  However, it is worth noting that, in this example, the equal-cost constraint
  does not explicitly match either of the error types, and therefore the two
  groups will in expectation experience different types of errors.
  In the left plot of \autoref{fig:health-res} (\hardtetal{}), we see that it is
  alternatively feasible to explicitly match both the F.P. and F.N. rates while sacrificing
  calibration.

\paragraph{Criminal Recidivism Prediction.}
  Finally, we examine the frameworks in the context of our motivating example: criminal recidivism.
  As mentioned in the introduction, African Americans ($G_1$) receive a disproportionate
  number of F.P. predictions as compared with Caucasians ($G_2$) when automated risk
  tools are used in practice.
  Therefore, we aim to equalize the generalized F.P. rate.
  In this experiment, we modify the predictions made by the COMPAS
  tool \cite{dieterich-northpointe-fairness}, a risk-assessment tool used in practice by the
  American legal system.
  Additionally, we also see if it is possible to improve the classifiers with training-time
  Equalized Odds constraints using the methods of \citet{zafar2017fairness} (\zafaretal{}).
  In \autoref{fig:compas-res}, we first observe that the original classifiers $h_1$ and $h_2$
  have large generalized F.P. and F.N. rates. Both methods of achieving
  Equalized Odds --- training constraints (left plot) and post-processing (middle plot)
  match the error rates while sacrificing calibration.
  However, we observe that, assuming $h_1$ and $h_2$ cannot be improved, it is
  infeasible to achieve the calibrated relaxation (\autoref{fig:compas-res} right).
  This is an example where matching the F.P. rate of $h_1$ would require
  a classifier worse than the trivial classifier $h^{\br_2}$.
  This example therefore represents an instance in which calibration is completely
  incompatible with any error-rate constraints.
  If the primary concern of criminal justice practitioners is calibration
  \cite{dieterich-northpointe-fairness,flores-re-propublica-fair}, then there
  will inherently be discrimination in the form of F.P. and F.N. rates.
  However, if the Equalized Odds framework is adopted, the miscalibrated risk
  scores inherently cause discrimination to one group, as argued in the introduction.
  Therefore, the most meaningful change in such a setting
  would be an improvement to $h_2$ (the classifier for African Americans)
  either through the collection of more data or the use of more salient features.
  A reduction in overall error to the
  group with higher cost will naturally lead to less error-rate disparity.


\section{Discussion and Conclusion}

  We have observed cases in which calibration and relaxed Equalized Odds
  are compatible and cases where they
  are not. When it is feasible, the penalty of equalizing cost is amplified if
  the base rates between groups differ significantly. This is expected,
  as base rate differences are what give rise to cost-disparity in the calibrated setting.
  Seeking equality with respect to a single error rate (e.g. false-negatives, as in the income
  prediction experiment) will necessarily increase disparity with respect to the
  other error. This may be tolerable (in the income prediction case, some
  employees will end up over-paid) but could also be highly problematic (e.g.
  in criminal justice settings).
%
%
  Finally, we have observed that the calibrated relaxation is infeasible when the best (discriminatory)
  classifiers are not far from the trivial classifiers (leaving little room for interpolation). In such settings,
  we see that calibration is completely incompatible with an equalized error
  constraint.

In summary, we conclude that maintaining cost parity \emph{and} calibration is
desirable yet often difficult in practice. Although we provide an algorithm to
effectively find the unique feasible solution to both constraints, it is
inherently based on randomly exchanging the predictions of the better
classifier with the trivial base rate. Even if fairness is reached in
expectation, for an individual case, it may be hard  to accept that
occasionally consequential decisions are made by randomly withholding
predictive information, irrespective of a particular person's feature representation. In this
paper we argue that, as long as calibration is required, no lower-error solution can be achieved.

\section*{Acknowledgements}
GP, FW, and KQW are supported in part by grants from the National Science Foundation
(III-1149882, III-1525919, III-1550179, III-1618134, and III-1740822),
the Office of Naval Research DOD (N00014-17-1-2175),
and the Bill and Melinda Gates Foundation.
MR is supported by an NSF Graduate Research Fellowship (DGE-1650441).
JK is supported in part by a Simons Investigator Award, an ARO MURI grant, a
Google Research Grant, and a Facebook Faculty Research Grant.

{\footnotesize
  \bibliographystyle{abbrvnat}
  \bibliography{citations}
}

%
%

\clearpage


\makeatletter
  \setcounter{table}{0}
  \renewcommand{\thetable}{S\arabic{table}}%
  \setcounter{figure}{0}
  \renewcommand{\thefigure}{S\arabic{figure}}%
  \setcounter{section}{0}
  \renewcommand{\thesection}{S\arabic{section}}
  \renewcommand{\thesubsection}{\thesection.\arabic{subsection}}
  \setcounter{equation}{0}
  \renewcommand\theequation{S\arabic{equation}}
  \renewcommand{\bibnumfmt}[1]{[S#1]}

  \newcommand{\suptitle}{Supplementary Information for: \titl}
  \renewcommand{\@title}{\suptitle}
  \newcommand{\thanks}[1]{\footnotemark[1]}
  \renewcommand{\@author}{\authorinfo}

  \par
  \begingroup
    \renewcommand{\thefootnote}{\fnsymbol{footnote}}
    \renewcommand{\@makefnmark}{\hbox to \z@{$^{\@thefnmark}$\hss}}
    \renewcommand{\@makefntext}[1]{%
      \parindent 1em\noindent
      \hbox to 1.8em{\hss $\m@th ^{\@thefnmark}$}#1
    }
    \thispagestyle{empty}
    \@maketitle
    \@thanks
  \endgroup
  \let\maketitle\relax
  \let\thanks\relax
\makeatother

For simplicity, our focus thus far has been on classifiers that are perfectly
calibrated. Here, we introduce an approximate notion of calibration, which we
will use in subsequent proofs.
\begin{definition}
  \label{def:calibration}
  The \emph{calibration gap} $\calib{h_t}$ of a classifier $h_t$
  with respect to a group $G_t$ is
  \begin{align}
    \calib{h_t}
    &= \int_{0}^{1} \Bigl\lvert \Pr_{(\x, y) \sim G_t} \bigl[ y\!=\!1 \mid h(\x)\!=\!p
  \bigr] - p \Bigr\rvert \Pr_{(\x, y) \sim G_t} \bigl[ h(\x) = p \bigr] dp.
      \label{eqn:approx-cal}
  \end{align}
  Thus, a classifier $h_t$ is \emph{perfectly calibrated} if $\calib{h_t} = 0$.
\end{definition}
A majority of this supplementary material is devoted to proving approximate
versions of our major findings. In all cases, our results degrade smoothly as
the calibration condition is relaxed.
In addition, we also provide extended details on the experiments run in this paper.

Note that we will use the notational abuse
$\Pr_{G_t}$ and $\ev_{G_t}$ in place of $\Pr_{(\x, y) \sim G_t}$
and $\ev_{(\x, y) \sim G_t}$.

\section{Linearity of Calibrated Classifiers}
\label{app:linearity}

  In \autoref{sec:setup}, we claim that the set of all calibrated classifiers
  $\mathcal{H}_t^*$ for group $G_t$ form a line in the generalized false-positive/false-negative
  plane. The following proof of this claim is adapted from \cite{kleinberg2016inherent}.
\begin{lemma}
  For a group $G_t$, if a classifier $h_t$ has $\epsilon(h_t) \le
  \delta_{cal}$, then
  \begin{align*}
    \bigl| \mu_t \fn{h_t} - \left( 1 - \mu_t \right) \fp{h_t} \bigr|
    \leq 2 \delta_{cal}.
  \end{align*}
  where $\fp{h_t}$ and $\fn{h_t}$ are the generalized false-positive and false-negative
  and $\mu_t$ is the base rate of group $G_t$.
  \label{lem:approx-linear-cal}
\end{lemma}
\begin{proof}
  First, note that
  \begin{align*}
    \fp{h_t} &= \ev_{G_t} \bigl[ h_t(\x) \mid y \! = \! 0 \bigr]
    \\&= \int_{0}^1 p \: \Pr_{G_t} \left[ h_t(\x) \! = \! p \mid y \! = \! 0 \right] dp
    \\&= \int_{0}^1 p \: \frac{1 - \Pr_{G_t} \left[ y \! = \! 1 \mid h_t(\x) \! = \! p \right]}{1 - \Pr_{G_t} \left[ y \! = \! 1 \right]}
     \Pr_{G_t} \left[ h_t(\x) \! = \! p \right] dp
    \\&= \frac{1}{1-\mu_t} \int_{0}^1 p \: (1 - \Pr_{G_t} \left[ y \! = \! 1 \mid h_t(\x) \! = \! p \right])
     \Pr_{G_t} \left[ h_t(\x) \! = \! p \right] dp
    \numberthis \label{eqn:cfp-int}
  \end{align*}
  Next, observe that
  \begin{align*}
    \int_0^1 & p \cdot \Pr_{G_t} \left[ y \! = \! 1 \mid h_t(\x) \! = \! p \right] \cdot
    \Pr_{G_t} \left[ h_t(\x) \! = \! p \right] dp
    \\&= \int_0^1 p (p + \Pr_{G_t} \left[ y \! = \! 1 \mid h_t(\x) \! = \! p \right] -
    p)
    \Pr_{G_t} \left[ h_t(\x) \! = \! p \right] dp
    \\&\le \int_0^1 \left(p^2 + |\Pr_{G_t} \left[ y \! = \! 1 \mid h_t(\x) \! = \! p \right] -
  p|\right)
  \Pr_{G_t} \left[ h_t(\x) \! = \! p \right] dp
    \\&\le \ev_{G_t}[h_t(\x)^2] + \delta_{cal}
  \end{align*}
  Similarly,
  \begin{align*}
    \int_0^1 & p \cdot \Pr_{G_t} \left[ y \! = \! 1 \mid h_t(\x) \! = \! p \right]
    \Pr_{G_t} \left[ h_t(\x) \! = \! p \right] dp
    \\&\ge \int_0^1 \left(p^2 - |\Pr_{G_t} \left[ y \! = \! 1 \mid h_t(\x) \! = \! p \right] -
  p|\right)
  \Pr_{G_t} \left[ h_t(\x) \! = \! p \right] dp
    \\&\ge \ev_{G_t}[h_t(\x)^2] - \delta_{cal}
  \end{align*}
  Plugging these into \eqref{eqn:cfp-int}, we have
  \begin{align*}
    \frac{1}{1-\mu_t} &(\ev_{G_t}[h_t(\x)] - \ev_{G_t}[h_t(\x)^2] -
    \delta_{cal}) \le \fp{h_t}
    \\&\le \frac{1}{1-\mu_t} (\ev_{G_t}[h_t(\x)] - \ev_{G_t}[h_t(\x)^2] +
    \delta_{cal}) \numberthis \label{eqn:fpbound}
  \end{align*}
  We follow a similar procedure for $\fn{h_t}$:
  \begin{align*}
    \fn{h_t} &= \ev_{G_t} \bigl[ 1 - h_t(\x) \mid y \! = \!0 \bigr]
    \\&= \int_{0}^1 \left( 1 - p \right) \: \Pr_{G_t} \left[ h_t(\x) \! = \! p \mid y \! = \! 0 \right] dp
    \\&= \int_{0}^1 \left( 1 - p \right) \: \frac{\Pr_{G_t} \left[ y \! = \! 1 \mid h_t(\x) \! = \! p \right]}{\Pr_{G_t} \left[ y \! = \! 1 \right]}
     \Pr_{G_t} \left[ h_t(\x) \! = \! p \right] dp.
    \\&= \frac{1}{\mu_t} \int_{0}^1 \left( 1 - p \right) (\Pr_{G_t} \left[ y \! = \! 1 \mid h_t(\x) \! = \! p \right])
    \Pr_{G_t} \left[ h_t(\x) \! = \! p \right] dp.
  \end{align*}
  We use the fact that
  \begin{align*}
    (1 - p)& (\Pr_{G_t} \left[ y \! = \! 1 \mid h_t(\x) \! = \! p \right])
    \\&=(1 - p) (p + \Pr_{G_t} \left[ y \! = \! 1 \mid h_t(\x) \! = \! p \right] - p)
    \\&\le
    p(1 - p) + |\Pr_{G_t} \left[ y \! = \! 1 \mid h_t(\x) \! = \! p \right] - p|
  \end{align*}
  and
  \begin{align*}
    (1 - p)& (\Pr_{G_t} \left[ y \! = \! 1 \mid h_t(\x) \! = \! p \right])
    \\&\ge
    p(1 - p) - |\Pr_{G_t} \left[ y \! = \! 1 \mid h_t(\x) \! = \! p \right] - p|
  \end{align*}
  to get
  \begin{align*}
    \frac{1}{\mu_t} &(\ev_{G_t}[h_t(\x)] - \ev_{G_t}[h_t(\x)^2] -
    \delta_{cal}) \le \fn{h_t}
    \\&\le \frac{1}{\mu_t} (\ev_{G_t}[h_t(\x)] - \ev_{G_t}[h_t(\x)^2] +
    \delta_{cal}) \numberthis \label{eqn:fnbound}
  \end{align*}
  Combining \eqref{eqn:fpbound} and \eqref{eqn:fnbound}, we have
  \begin{align*}
    \fn{h_t} &\le
    \frac{1}{\mu_t} (\ev_{G_t}[h_t(\x)] - \ev_{G_t}[h_t(\x)^2] +
    \delta_{cal})
    \\&= \frac{1}{\mu_t} (\ev_{G_t}[h_t(\x)] - \ev_{G_t}[h_t(\x)^2] -
    \delta_{cal} + 2\delta_{cal})
    \\&\le \frac{1}{\mu_t} ( \left(1 - \mu_t\right) \fp{h_t} + 2\delta_{cal})
    \\&= \frac{1 - \mu_t}{\mu_t} \fp{h_t} + \frac{2\delta_{cal}}{\mu_t}
  \end{align*}
  We can get a similar lower bound for \fn{h_t} as
  \begin{align*}
    \fn{h_t} &\ge
    \frac{1}{\mu_t} (\ev_{G_t}[h_t(\x)] - \ev_{G_t}[h_t(\x)^2] -
    \delta_{cal})
    \\&\ge \frac{1}{\mu_t} (\left( 1 - \mu_t \right) \fp{h_t} - 2\delta_{cal})
    \\&= \frac{1 - \mu_t}{\mu_t} \fp{h_t} - \frac{2\delta_{cal}}{\mu_t}
  \end{align*}
  Multiplying these inequalities by $\mu_t$ completes this proof.
\end{proof}
\begin{corollary}
  Let $\mathcal H_t$ be the set of perfectly calibrated classifiers
  for group $G_t$ --- i.e. for any $h^*_t \in \mathcal H_T$, we have
  $\calib{h^*_t} = 0$.
  The generalized false-positive and false-negative rates of
  $h^*_t$ are given by
  \begin{align}
    \fp{h^*_t} &= \frac{1}{1 - \mu_t} \left( \ev_{G_t}[h_t(\x)] - \ev_{G_t}[h_t(\x)^2] \right)
      \label{eqn:fp_exact} \\
    \fn{h^*_t} &= \frac{1}{\mu_t} \left( \ev_{G_t}[h_t(\x)] - \ev_{G_t}[h_t(\x)^2] \right)
      \label{eqn:fn_exact}
  \end{align}
\end{corollary}
\begin{proof}
  This is a direct consequence of \eqref{eqn:fpbound} and \eqref{eqn:fnbound}.
\end{proof}
\begin{corollary}
  \label{crly:linearity}
  For a group $G_t$, any perfectly calibrated classifier $h^*_t$ satisfies
  \begin{equation}
    \fn{h^*_t} = \frac{1 - \mu_t}{\mu_t} \fp{h_t}.
    \label{eqn:linear-calib-sup}
  \end{equation}
\end{corollary}
In other words, all perfectly calibrated classifiers $h^*_t \in \mathcal H_t$ for group $G_t$
lie on a line in the generalized false-positive/false-negative plane, where the slope of the
line is uniquely determined by the group's base-rate $\br_t$.

\section{Cost Functions}
\label{app:cost-funcs}

We will prove a few claims about cost functions
$\fairconstname$ of the form given by \eqref{eqn:cost_general} ---
i.e.
\[ \fairconst{h_t}{t} = a_t \fp{h_t} + b_t \fn{h_t} \]
for some non-negative constants $a_t$ and $b_t$.
First, we show that $h^{\br_t}$ is the calibrated classifier that maximizes
$\fairconstname$.


\begin{lemma}
  For any cost function $\fairconstname$ that follows the form of
  \eqref{eqn:cost_general}, the trivial classifier $h^{\br_t}$ is the calibrated
  classifier for $G_t$ with maximum cost.
  \label{lma:trivial}
\end{lemma}
\begin{proof}
  Again, let $\fairconstname$ be a cost function:
  $$\fairconst{h}{t} = a_t \fp{h_t} + b_t \fn{h_t}.$$
  Using \eqref{eqn:fp_exact} and \eqref{eqn:fn_exact}, we have that, for every classifier $h_t$
  that is perfectly calibrated for group $G_t$,
  \begin{align*}
    \fairconst{h_t}{t} &= a_t \fp{h_t} + b_t \fn{h_t}
    \\&= \left( \frac{a_t}{1 - \mu_t} + \frac{b_t}{\mu_t} \right) \left( \ev_{G_t} \left[ h_t(x) \right] - \ev_{G_t} \left[ h_t(x)^2 \right] \right)
    \\&= \left( \frac{a_t}{1 - \mu_t} + \frac{b_t}{\mu_t} \right) \left( \mu_t - \ev_{G_t} \left[ h_t(x)^2 \right] \right).
  \end{align*}
  The last equation holds because $\ev_{G_t} \left[ h_t(\x) \right] = \mu_t$ for any calibrated
  classifier -- a fact which can easily be derived from \autoref{def:calibration}.

  We would like to find, $h^{\max}_t \in \mathcal{H}^*_t$, the calibrated classifier with the highest weighted cost.
  Because $\left( \frac{a_t}{1 - \mu_t} + \frac{b_t}{\mu_t} \right)$ and $\mu_t$ are non-negative constants, we have
  \begin{align*}
    h^{\max}_t &= \argmax_{h \in \mathcal{H}^*_t} \Bigl[ \left( \frac{a_t}{1 - \mu_t} + \frac{b_t}{\mu_t} \right) \left( \mu_t - \ev_{G_t} \left[ h(x)^2 \right] \right) \Bigr]
    \\&= \argmax_{h \in \mathcal{H}^*_t} \Bigl[ - \ev_{G_t} \left[ h(x)^2 \right] \Bigr]
    \\&= \argmin_{h \in \mathcal{H}^*_t} \Bigl[ \ev_{G_t} \left[ h(x)^2 \right] \Bigr]
    \\&= \argmin_{h \in \mathcal{H}^*_t} \Bigl[ \ev_{G_t} \left[ h(x)^2 \right] - \mu_t^2 \Bigr]
  \end{align*}
  Thus, the calibrated classifier with minimum variance will have the highest cost. This translates to
  a classifier that outputs the same probability for every sample. By the calibration constraint, this
  constant must be equal to $\mu_t$, so this classifier must be the trivial classifier $h^{\br_t}$ --- i.e.
  for all $\x$
  \[ h^{\max}_t \left( \x \right) = h^{\br_t} \left( \x \right) = \mu_t. \]
\end{proof}

Next, we show that $\fairconstname$ is linear under randomized interpolations.
\begin{lemma} \label{lma:interpolation}
  Let $\fair h_2$ be the classifier derived from \eqref{eqn:interpolation}
  with interpolation parameter $\alpha \in [0, 1]$.
  The cost of $\fair h_2$ is given by
  \[
    \fairconst{\fair h_2}{2} = (1 - \alpha) \fairconst{h_2}{2} + \alpha \fairconst{h^{\br_2}}{2}
  \]
\end{lemma}
\begin{proof}
  The cost of $\fair h_2$ can be calculated using linearity of expectation. Let
  $B$ be a Bernoulli random variable with parameter $\alpha$.
  \begin{align*}
    \fairconst{\fair h_2}{2} &= a_2 \fp{\fair h_2} + b_2 \fn{\fair h_2}
    \\&= a_2 \ev_{G_2} \left[ 1 -  \fair h_2(\x) \mid y\!=\!1 \right] + b_2 \ev_{G_2} \left[ \fair h_2(\x) \mid y\!=\!0 \right]
    \\&= a_2 \ev_{B,G_2} \left[ 1 - \left[ (1-B) h_2(\x) + B h^{\br_2}(\x) \right] \mid y\!=\!1 \right]
    + b_2 \ev_{B,G_2} \left[ \left[ (1-B) h_2(\x) + B h^{\br_2}(\x) \right] \mid y\!=\!0 \right]
    \\&=                a_2 \ev_{B,G_2} \left[ (1-B) \left( 1 - h_2(\x) \right) \mid y\!=\!1 \right]
    + a_2 \ev_{B,G_2} \left[ B \left( 1 -  h^{\br_2}(\x) \right) \mid y\!=\!1 \right]
    \\&\phantom{=} \:
    + b_2 \ev_{B,G_2} \left[ (1-B) h_2(\x) \mid y\!=\!0 \right]
    + b_2 \ev_{B,G_2} \left[ B h^{\br_2}(\x) \mid y\!=\!0 \right]
    \\&=                a_2 \ev_{B} \left[ 1-B \right] \ev_{G_2} \left[ 1 - h_2(\x) \mid y\!=\!1 \right]
    + a_2 \ev_{B} \left[ B \right]   \ev_{G_2} \left[ 1 -  h^{\br_2}(\x) \mid y\!=\!1 \right]
    \\&\phantom{=} \:
    + b_2 \ev_{B} \left[ 1-B \right] \ev_{G_2} \left[ h_2(\x) \mid y\!=\!0 \right]
    + b_2 \ev_{B} \left[ B \right]   \ev_{G_2} \left[ h^{\br_2}(\x) \mid y\!=\!0 \right]
    \\&= a_2 (1-\alpha) \fp{h_2} + b_2 (1-\alpha) \fn{h_2}
    + a_2 (\alpha) \fp{h^{\br_2}} + b_2 (\alpha) \fn{h^{\br_2}}
    \\&= (1 - \alpha) \fairconst{h_2}{2} + \alpha \fairconst{h^{\br_2}}{2}.
  \end{align*}
\end{proof}

\section{Relationship Between Cost and Error}
\label{sec:cost_and_error}

In \autoref{sec:setup}, we claim that there is a tight connection
between reducing any cost function $\fairconst{h_t}{t}$ and reducing the generalized
error rates $\fp{h_t}$ and $\fn{h_t}$
for approximately calibrated classifiers.
In other words, assuming we are approximately calibrated, improving cost will
approximately improve our error rates.
We formalize this notion in this section:
\begin{lemma}
  \label{lma:cost_error}
  Let $h_t$ be a classifier with $\epsilon(h_t) = \delta_{cal}$ and cost
  $\fairconst{h_t}{t}$. For any other classifier $h_t'$, if $\fp{h_t'} <
  \fp{h_t} - \frac{4\delta_{cal}}{1-\mu_t}$ or $\fn{h_t'} < \fn{h_t} -
  \frac{4\delta_{cal}}{\mu_t}$, then $\fairconst{h_t'}{t} <
  \fairconst{h_t}{t}$ or $\epsilon(h_t) > \delta_{cal}$.
  \label{lem:fpfn}
\end{lemma}
\begin{proof}
  First, assume that $\fp{h_t'} < \fp{h_t} - \frac{4\delta_{cal}}{1-\mu_t}$.
  Then, there are two cases: either $\fn{h_t'} < \fn{h_t}$ or $\fn{h_t'} \ge
  \fn{h_t}$. In the first case, $\fairconst{h_t'}{t} < \fairconst{h_t}{t}$
  because $\fp{h_t'} < \fp{h_t}$ and $\fn{h_t'} < \fn{h_t}$. In the second case,
  if $\epsilon(h_t') \le \delta_{cal}$, we can use Lemma
  \ref{lem:approx-linear-cal} to get
  \begin{align*}
    \fp{h_t'}
    &\ge \frac{\mu_t}{1-\mu_t} \fn{h_t'} -
    \frac{2\delta_{cal}}{1-\mu_t} \\
    &\ge \frac{\mu_t}{1-\mu_t} \fn{h_t} -
    \frac{2\delta_{cal}}{1-\mu_t} \\
    &\ge \fp{h_t} -
    \frac{4\delta_{cal}}{1-\mu_t}
  \end{align*}
  Since this contradicts the initial assumption that $\fp{h_t'} < \fp{h_t} -
  \frac{4\delta_{cal}}{1-\mu_t}$, it cannot be the case that $\epsilon(h_t') \le
  \delta_{cal}$. This proves the lemma when $\fp{h_t'} < \fp{h_t} -
  \frac{4\delta_{cal}}{1-\mu_t}$.

  To prove the second part, we now assume that $\fn{h_t'} < \fn{h_t} -
  \frac{4\delta_{cal}}{\mu_t}$. We again break this into two cases. If
  $\fp{h_t'} < \fp{h_t}$, then $\fairconst{h_t'}{t} < \fairconst{h_t}{t}$. If
  $\fp{h_t'} \ge \fp{h_t}$, then under the assumption that $\epsilon(h_t') \le
  \delta_{cal}$, we can rearrange Lemma \ref{lem:approx-linear-cal} to get
  \begin{align*}
    \fn{h_t'}
    &\ge \frac{1-\mu_t}{\mu_t} \fp{h_t'} -
    \frac{2\delta_{cal}}{\mu_t} \\
    &\ge \frac{1-\mu_t}{\mu_t} \fp{h_t} -
    \frac{2\delta_{cal}}{\mu_t} \\
    &\ge \fn{h_t} -
    \frac{4\delta_{cal}}{\mu_t}
  \end{align*}
  Again, this contradicts the assumption that $\fn{h_t'} < \fn{h_t} -
  \frac{4\delta_{cal}}{\mu_t}$, so it cannot be the case that
  $\epsilon(h_t') \le \delta_{cal}$. This completes the proof.
\end{proof}
From this result we can derive a stronger claim
for perfectly calibrated classifiers.
\begin{lemma}
  Let $h_t$ and $h'_t$ be perfectly calibrated classifiers with cost
  $\fairconst{h_t}{t} \leq \fairconst{h'_t}{t}$ for some
  cost function $\fairconstname$. Then $\fp{h_t} \leq \fp{h'_t}$
  and $\fn{h_t} \leq \fn{h'_t}$, with equality only if $\fairconst{h_t}{t} = \fairconst{h'_t}{t}$.
\end{lemma}

\section{Proof of \autoref{alg:method} Optimality and Approximate Optimality}
\label{sec:optimality-proofs}

In \autoref{sec:relaxation}, we claim that \autoref{alg:method} produces
optimal non-discriminatory classifiers in exact calibration scenarios,
and near-optimal classifiers in approximate calibration scenarios.

We begin with classifiers $h_1$ and $h_2$ be classifiers for groups $G_1$ and
$G_2$, with calibrations $\epsilon(h_1) \le \delta_{cal}$ and $\epsilon(h_2)
\le \delta_{cal}$. As before, assume that we cannot strictly
improve the cost of either $h_1$ or $h_2$ without worsening calibration: i.e. $h_1$ and
$h_2$ is $\fairconstname$ and calibration. We will now show that
\autoref{alg:method} produces classifiers that are
near-optimal with respect to both the false-positive and false-negative rates
among calibrated classifiers satisfying the equal-cost constraint.

First, we show that interpolation preserves approximate calibration:
\begin{thm}[Approximate Optimality of \autoref{alg:method}]
  Given $\fair h_2$, which is the classifier produced by \autoref{alg:method},
  we have that $\calib{\fair h_2} \leq \left( 1 - \alpha \right) \calib{h_2}$, where
  $\alpha \in [0, 1]$ is the interpolation parameter in \eqref{eqn:interpolation}.
  \label{thm:calibration}
\end{thm}
\begin{proof}
  We can calculate the calibration of $\fair h_2(\x)$ as follows:
  \begin{align*}
    \calib{\fair h_2}
    &= \mathop{\ev}_{B, G_2} \Bigl\lvert \Pr_{G_2} \bigl[ y\!=\!1 \mid \fair
    h_2(\x)\!=\!p \bigr] - p \Bigr\rvert
    \\
    &= \int_{0}^1 \Bigl\lvert \Pr_{G_2} \bigl[ y\!=\!1 \mid \fair
    h_2(\x) \!=\!p \bigr] - p \Bigr\rvert
    \Pr_{G_2} \bigl[ \fair h_2(\x)\!=\!p\bigr] dp
  \end{align*}
  For $p \ne \mu_2$, $|\Pr_{G_2} \bigl[ y\!=\!1 \mid \fair h_2(\x) \!=\!p \bigr]
  - p| = |\Pr_{G_2} \bigl[ y\!=\!1 \mid h_2(\x) \!=\!p \bigr] - p|$ and
  $\Pr_{G_2} \bigl[ \fair h_2(\x)\!=\!p\bigr] = (1-\alpha)\Pr_{G_2} \bigl[
  h_2(\x)\!=\!p\bigr]$.

  For $p = \mu_2$, let $\beta = \Pr_{B,G_2}\bigl[B \!=\! 1 \mid \fair h_2(\x) =
  p\bigr]/\Pr_{G_2} \bigl[\fair h_2(\x) = p\bigr]$. Note that $\beta \ge
  \alpha$. Then,
  \begin{align*}
    \Pr_{G_2} \bigl[ y\!=\!1 \mid \fair h_2(\x) \!=\!p \bigr] &=
    (1-\beta) \Pr_{G_2} \bigl[ y\!=\!1 \mid h_2(\x) \!=\!p \bigr] + \beta
    \Pr_{G_2} \bigl[ y\!=\!1 \mid h^{\br_2}(\x) \!=\!p \bigr] \\
    &= (1-\beta) \Pr_{G_2} \bigl[ y\!=\!1 \mid h_2(\x) \!=\!p \bigr] + \beta p
  \end{align*}
  because $h^{\br_2}$ is perfectly calibrated. Moreover, note that $\Pr_{G_2}
  \bigl[ \fair h_2(\x)\!=\!p\bigr] = \Pr_{G_2} \bigl[ h_2(\x)\!=\!p\bigr] /
  (1-\beta)$
  Using this, we have
  $|\Pr_{G_2} \bigl[ y\!=\!1 \mid \fair h_2(\x) \!=\!p
  \bigr] - p | \Pr_{G_2} \bigl[ \fair h_2(\x)\!=\!p\bigr] =
  |\Pr_{G_2} \bigl[ y\!=\!1 \mid h_2(\x) \!=\!p
  \bigr] - p | \Pr_{G_2} \bigl[ h_2(\x)\!=\!p\bigr]$.
  Thus,
  \begin{align*}
    \calib{\fair h_2}
    &= \int_{0}^1 \Bigl\lvert \Pr_{G_2} \bigl[ y\!=\!1 \mid \fair
    h_2(\x) \!=\!p \bigr] - p \Bigr\rvert
    \Pr_{G_2} \bigl[ \fair h_2(\x)\!=\!p\bigr] dp \\
    &\le \int_{0}^1 \Bigl\lvert \Pr_{G_2} \bigl[ y\!=\!1 \mid
    h_2(\x) \!=\!p \bigr] - p \Bigr\rvert
    \Pr_{G_2} \bigl[ h_2(\x)\!=\!p\bigr] dp \\
    &= \calib{h_2}
  \end{align*}
\end{proof}

Next, we observe that by \autoref{lma:cost_error}, for any
classifiers $h_1'$ and $h_2'$ with $\epsilon(h_1') \le \delta_{cal}$ and
$\epsilon(h_2') \le \delta_{cal}$ satisfying the equal-cost constraint, it must be
the case that $\fp{h_t'} \ge \fp{\fair h_t} - \frac{4 \delta_{cal}}{1-\mu_t}$
and $\fn{h_t'} \ge \fp{\fair h_t} - \frac{4 \delta_{cal}}{\mu_t}$ for $t = 1, 2$.

Thus, approximately calibrated classifiers will be approximately optimal. From
this result, it is easy to derive the optimality result for perfectly-calibrated
classifiers.
\begin{thm}[Exact Optimality of \autoref{alg:method}]
  \label{thm:optimality}
  \autoref{alg:method} produces the classifiers $h_1$ and $\fair h_2$ that
  satisfy both perfect calibration and the equal-cost constraint with the lowest
  possible generalized false positive and false negative rates.
\end{thm}

\section{Proof of Impossibility and Approximate Impossibility}
\label{app:impossiblility}

In this section, we prove that it is impossible to satisfy multiple
equal-cost constraints while simultaneously satisfying calibration.
We will first prove this in an exact sense, and then show that
the result holds approximately as well.

\subsection{Exact Impossibility Theorem}

\newtheorem*{thm:impossibility}{Theorem \ref{thm:impossibility} (Restated)}
\begin{thm:impossibility}
  Let $h_1$ and $h_2$ be calibrated classifiers for $G_1$ and $G_2$ with equal
  cost with respect to
  $\fairconstname$.
  If $\mu_1 \ne \mu_2$, and if $h_1$ and $h_2$ have equal cost with respect to
  $\fairconstname'$, then $h_1$ and $h_2$ must be perfect classifiers.
\end{thm:impossibility}
\begin{proof}
  First, observe that the perfect classifier always satisfies any equal-cost
  constraint simply because if $\fp{t} = \fn{t} = 0$,
  $\fairconst{h_t}{t} = 0$. Moreover, the perfect classifier is
  always calibrated.

  For any classifier, as shown in \cite{kleinberg2016inherent}, \fp{h_t} and
  \fn{h_t} are linearly related by \eqref{eqn:linear-calib-sup}.
  Furthermore, each equal-cost constraint is linear in \fp{h_t} and
  \fn{h_t}. We define \fairconst{h_t}{t} and
  \fairconstpr{h_t}{t} to be identical cost functions if
  the equal-cost constraints that they impose are identical, meaning one
  constraint is satisfied if and only if the other is satisfied. If this is not
  the case, then \fairconst{h_t}{t} and
  \fairconstpr{h_t}{t} are \emph{distinct},
  meaning that
  the equal-cost constraints are linearly independent for $\mu_1 \ne \mu_2$.
  Moreover, these are also linearly independent from the calibration constraints
  because by assumption, they both have nonzero coefficients for at least one of
  $(\fp{h_1}, \fn{h_1})$ and $(\fp{h_2}, \fn{h_2})$. As a result, we have four linearly
  independent constraints (2 from calibration and at least 2 equal-cost
  constraints) on 4 variables (\fp{h_1}, \fn{h_1}, \fp{h_2}, \fn{h_2}), meaning that these
  constraints yield a unique solution. From above, we know that all the
  constraints
  are simultaneously satisfied when $\fp{h_t} = \fn{h_t} = 0$ for $t = 1,2$, meaning
  that the perfect classifier is the only classifier for which they are
  simultaneously satisfied.
\end{proof}

\subsection{Approximate Impossibility Theorem}

Now, we will show that this impossibility result holds in an approximate sense --- i.e.
approximately satisfying the calibration and equal-cost constraints is only possible
if the classifiers approximately perfect.

Since the calibration and equal-cost constraints are all linear, let $A$ be the
matrix that encodes them. With two equal-cost constraints
\fairconstname{} and
$\fairconstname'$,
\[
  A = \begin{bmatrix}
    1 &-\frac{\mu_1}{1-\mu_1} & 0 & 0 \\
    0 & 0 & 1 &-\frac{\mu_2}{1-\mu_2} \\
    a_1 & b_1 & -a_2 & -b_2 \\
    a_1' & b_1' & -a_2' & -b_2'
  \end{bmatrix}.
\]
Note that the first two rows of $A$ encode the calibration conditions --- see \eqref{eqn:linear-calib-sup}.
The bottom two rows encode two equal-cost constraints.
Furthermore, let
\[
  \vec{q} = \left[\fp{h_1} \; \fn{h_1} \; \fp{h_2} \; \fn{h_2}\right]^\top.
\]
If all constraints are required to hold exactly, then we have $A \vec{q} = 0$.
Consider the case where the calibration and equal-cost constraints hold
approximately.
\begin{thm}[Generalized approximate impossibility result] \label{thm:approx-impossibility}
  Let $h_1$ and $h_2$ be classifiers with calibration $\delta_{cal}$ and
  cost difference at most $\delta_{cost}$ with respect to distinct cost functions
  $\fairconstname$ and $\fairconstname'$. Furthermore, assume that every entry
  of $A$ is rational with some common denominator $D$ and is upper bounded by
  some maximum value $M$. Then, there is a constant $L$ that depends on $D$ and
  $M$ such that
  \[
    \fp{h_t} \le L \cdot \max\left\{\frac{2\delta_{cal}}{1-\mu_1},
    \frac{2\delta_{cal}}{1-\mu_2}, \delta_{cost}\right\}
  \]
  and
  \[
    \fn{h_t} \le L \cdot \max\left\{\frac{2\delta_{cal}}{1-\mu_1},
    \frac{2\delta_{cal}}{1-\mu_2}, \delta_{cost}\right\}
  \]
  for $t = 1,2$.
  \label{thm:approx}
\end{thm}

\newcommand{\wA}{\widehat{A}}
\begin{proof}
  By Lemma \ref{lem:approx-linear-cal},
  \[
    \bigl|\mu_t \fn{h_t} - \left(1 - \mu_t \right) \fp{h_t} \bigr| \le 2
      \delta_{cal}.
  \]
  Since the first two rows in $A$ correspond to the calibration constraints, and
  the second to correspond to the equal-cost constraints, it must be the case
  that
  \[
    |A \vec{q}| \le \begin{bmatrix}
      \frac{2\delta_{cal}}{1-\mu_1} \\ \frac{2 \delta_{cal}}{1-\mu_2} \\
      \delta_{cost} \\ \delta_{cost}
    \end{bmatrix},
  \]
  i.e. the absolute value of each entry in $A\vec{q}$ is bounded by the vector
  on the right hand side. Let $\vec{\nu} = [ 2\delta_{cal}/(1-\mu_1) \; 2
  \delta_{cal}/(1-\mu_2) \; \delta_{cost} \; \delta_{cost}]^\top$.
  Let $\vec{s} = sign(A \vec{q})$, and multiply the
  $i$th row of $A$ by the $i$th entry of $\vec{s}$ to produce $\wA$
  This allows us to drop the absolute value, meaning we have
  \[
    \wA \vec{q} \le \vec{\nu}
  \]
  Furthermore, since $\fairconstname$ and $\fairconstname'$ were assumed to be
  distinct, $\wA$ is invertible, so this is equivalent to
  \[
    \vec{q} \le \wA^{-1}\vec{\nu}.
  \]
  Taking $\ell_{\infty}$ norms of both sides,
  \[
    \|\vec{q}\|_\infty \le \|\wA^{-1} \vec{\nu}\|_\infty \le \|
    \wA^{-1}\|_\infty \|\vec{\nu}\|_\infty.
  \]
  The $(i,j)$ entry of $\wA^{-1}$ can be expresed as $\wA_{ji}/\det(\wA)$,
  where $\wA_{ji}$ is the $(j,i)$ cofactor. Note that $\wA_{ji}$ is a $3 \times 3$
  determinant, so it is the sum of $6$ cubic polynomials in entries of $\wA$.
  However, since every $3 \times 3$ submatrix of $\wA$ has at least one $0$
  entry, only $4$ of those cubics can be nonnegative. By assumption, the maximum
  value of any entry of $\wA$ is $M$, so $|\wA_{ji}| \le 4 M^3$.

  We can lower bound $\det(\wA)$ by noting that since $\wA$ is not singular, its
  determinant is nonzero. However, because the determinant can be expressed as a
  $4 \times 4$ polynomial, and each term has common denominator $D$ by
  assumption, $|\det(\wA)| \ge 1/D^4$. As a result, $|\wA_{ji}/\det(\wA)| \le 4
  M^3 D^4$.

  Let $d_{ij}$ be the $(i,j)$ entry of $\wA$. We know that
  \[
    \|\wA^{-1}\|_\infty \le \max_j \sum_{i=1}^4 |d_{ij}| \le 16M^3 D^4 = L
  \]
  As a result,
  \[
    \|\vec{q}\|_\infty \le L \|\nu\|_\infty
  \]
  which proves the claim.
\end{proof}
Note that \autoref{thm:approx} is not intended to be a tight bound. It simply
shows that impossibility result degrades smoothly for approximate constraints.

\section{Details on Experiments}
\label{app:experiments}

\paragraph{Post-processing for Equalized Odds}
To derive classifiers that satisfy the Equalized Odds notion of
fairness, we use the method introduced by \citet{hardt2016equality}.
Essentially, the false-positive and false-negative constraints
are satisfied by randomly flipping some of the predictions of the
original classifiers. Let $q^{(t)}_\text{n2p}$ be the probability for group $G_t$ of
``flipping'' a negative prediction to positive,
and $q^{(t)}_\text{p2n}$ be that of flipping a positive prediction to negative.
The derived classifiers $\eo h_1$ and $\eo h_2$ essentially flip predictions according to
these probabilities:
\begin{align*}
  \eo h_t(\x) &= \begin{cases}
    \left( 1 - h_t(\x) \right) B^{(t)}_\text{p2n}
    + h_t(\x) \left(1 - B^{(t)}_\text{p2n} \right)
    & h_t(\x) \geq 0.5
    \\
    \left( 1 - h_t(\x) \right) B^{(t)}_\text{n2p}
    + h_t(\x) \left(1 - B^{(t)}_\text{n2p} \right)
    & h_t(\x) < 0.5
  \end{cases}
\end{align*}
where $B^{(t)}_\text{n2p}$ and $B^{(t)}_\text{p2n}$ are Bernoulli random variables
with expectations $q^{(t)}_\text{n2p}$ and $q^{(t)}_\text{p2n}$ respectively.
Note that this is a probabilistic generalization of the derived classifiers
presented in \cite{hardt2016equality}. If all outputs of $h_t$ were either
$0$ or $1$ we would arrive at the original formulation.

We can find the best rates $q^{(1)}_\text{n2p}$, $q^{(1)}_\text{p2n}$,
$q^{(2)}_\text{n2p}$, and $q^{(2)}_\text{p2n}$ through the following optimization
problem:
\begin{mini*}
{q^{(1)}_\text{}, q^{(1)}_\text{p2n}, q^{(2)}_\text{n2p}, q^{(2)}_\text{p2n}}
{\mathcal{L}(\eo h_1) + \mathcal{L}(\eo h_2)}
{}{}
\addConstraint{\fp{\eo h_1} = \fn{\eo h_1}}
\addConstraint{\fp{\eo h_2} = \fn{\eo h_2}}
\end{mini*}
where $\mathcal{L}$ represents the $0$/$1$ loss of the classifier:
\begin{align*}
  \mathcal L (h_t) &=
  \Pr_{G_t} \left[ h_t(\x) \geq 0.5 \mid y \!=\! 0 \right]
  +
  \Pr_{G_t} \left[ h_t(\x) < 0.5 \mid y \!=\! 1 \right].
\end{align*}
The two constraints
enforce the Equalized Odds constraints.
\citet{hardt2016equality} show that this can be solved via
a linear program.

\paragraph{Constrained-learning for Equalized Odds}
\citet{zafar2017fairness} introduce a method to achieve Equalized Odds
(under the name \emph{Disparate Mistreatment}) at training time using
optimization constraints. The problem is set up at learning a logistic
classifier under the Equalized Odds constraints. While these constraints
make the problem non-convex, \citeauthor{zafar2017fairness} show how
to formulate the problem as a disciplined convex-concave program.
Though this is generally intractable, it can be solved in many instances.
We refer the reader to \cite{zafar2017fairness} for details.

\paragraph{Training Procedure for Income Prediction.}
We train three models: a random forest, a multi-layer perceptron, and a SVM
with an RBF kernel. We convert the categorical features into one-hot enocodings.
$10\%$ of the data is reserved for hyperparameter tuning and
post-processing, and an additional $10\%$ is saved for final evaluation.
The random forest and MLP are naturally probabilistic and well calibrated.
We use Platt scaling to calibrate the SVM.
The hyperparameters were tuned by grid search based on 3-fold cross validation.
\autoref{fig:adult-res} displays the average false-positive and false-negative
costs across all models.

\paragraph{Training Procedure for Health Prediction.}
We train  a random forest and a linear SVM on this dataset. We use the
same dataset split and hyperparameter selection as with Income Prediction.
\autoref{fig:health-res} displays the average false-positive and false-negative
costs across all models.

\paragraph{Training Procedure for Health Prediction.}
For the trained Equalized Odds baseline we train a constrained
logistic classifier using the method proposed by \cite{zafar2017fairness}.
We derive the post-processed
classifiers (both for Equalized Odds and its calibrated relaxation) from the
original COMPAS classifier \cite{dieterich-northpointe-fairness}.

\end{document}